\documentclass[runningheads]{llncs}
\usepackage{graphicx}
\usepackage{svg}
\usepackage{amsmath}
\usepackage{amssymb}
\usepackage{easyReview}
\usepackage{multirow}
\usepackage{array}
\usepackage{dcolumn}
\usepackage{cite}
\usepackage{makecell}
\usepackage[T1]{fontenc}

\newcolumntype{P}[1]{>{\centering\arraybackslash}p{#1}}
\newcolumntype{C}[1]{>{\centering\arraybackslash}m{#1}}

\begin{document}
\title{Class-Based Time Series Data Augmentation to Mitigate Extreme Class Imbalance for Solar Flare Prediction}

\author{
Junzhi Wen\thanks{Corresponding author: Junzhi Wen\\jwen6@student.gsu.edu}\orcidID{0000-0002-9176-5273}
\and Rafal A. Angryk\orcidID{0000-0001-9598-8207}
}
\institute{Department of Computer Science\\ Georgia State University\\Atlanta, GA 30302, USA\\\email{jwen6@student.gsu.edu,angryk@cs.gsu.edu}}

%
%
%
\maketitle              
\begin{abstract}
Time series data plays a crucial role across various domains, making it valuable for decision-making and predictive modeling. Machine learning (ML) and deep learning (DL) have shown promise in this regard, yet their performance hinges on data quality and quantity, often constrained by data scarcity and class imbalance, particularly for rare events like solar flares. Data augmentation techniques offer a potential solution to address these challenges, yet their effectiveness on multivariate time series datasets remains underexplored. In this study, we propose a novel data augmentation method for time series data named Mean Gaussian Noise (MGN).  We investigate the performance of MGN compared to eight existing basic data augmentation methods on a multivariate time series dataset for solar flare prediction, SWAN-SF, using a ML algorithm for time series data, TimeSeriesSVC. The results demonstrate the efficacy of MGN and highlight its potential for improving classification performance in scenarios with extremely imbalanced data. Our time complexity analysis shows that MGN also has a competitive computational cost compared to the investigated alternative methods.

\end{abstract}
\section{Introduction}
Time series data, characterized by a sequence of observations recorded over time intervals, has been permeating numerous domains such as finance \cite{tsay2005analysis}, healthcare \cite{gao2023interpretable}, climate science \cite{mudelsee2019trend}, and industrial processes \cite{mehdiyev2017time}. Time series classification, the task of categorizing time series data instances into distinct categories, is of paramount importance as it facilitates decision-making, anomaly detection, and predictions across diverse applications. Multiple applications of machine learning (ML) have been build and deployed in these areas over recent years. However, the performance of ML models depends on the quality and quantity of the data available, which are often limited in real-world scenarios. Especially for rare events such as solar flares, hurricanes, and earthquakes, the data is not only scarce but also characterized by extreme imbalances. Class imbalance \cite{batista2004study} is one of the major challenges for machine learning-based classification tasks, since when the data is extremely imbalanced, it will often introduce bias toward the majority classes, yielding unsatisfactory performance of the classifications for the extremely infrequent (but often more valuable) events. 


Data augmentation as a tool to enhance size of training data for ML purposes has been growing in popularity in recent years and has shown its effectiveness in many applications, such as image classifications \cite{shorten2019survey} and natural language processing (NLP) \cite{shorten2021text}. In this study, we introduce a novel data augmentation approach named Mean Gaussian Noise (MGN), which deviates from existing methods by synthesizing the underrepresented class globally. To evaluate MGN, we compare it with eight established basic data augmentation methods that have been studied in \cite{iwana2021empirical}. While \cite{iwana2021empirical} has primarily used univariate time series data, our investigation focuses on multivariate time series data, specifically targeting solar flare prediction using SWAN-SF \cite{angryk2020multivariate}. We conduct an experiment to compare the performance of these methods on a classification task using a ML algorithm for time series data. Moreover, we analyze the time complexity of those methods in practical settings.

The rest of the paper is organized as follows:  Section \ref{section:background} summarizes the basic augmentation methods for time series data and the MVTS dataset used in this study. Section \ref{section:mgn} provides an detailed introduction to the Mean Gaussian Noise (MGN) method. Experiment setup details are presented in Section \ref{section:exp_setup}, followed by a discussion of results in Section \ref{section:exp_result}. Section \ref{section:time} experimentally demonstrates the time complexity of different methods in practice. Finally, conclusions and avenues for future research are presented in Section \ref{section:conclusion}.

\section{Background}\label{section:background}   
\subsection{Basic Data Augmentation Methods for Time Series Data}\label{section:basic_augmentation}
\begin{figure*}[]
        \centering
        \includegraphics[width=0.95\textwidth, keepaspectratio]{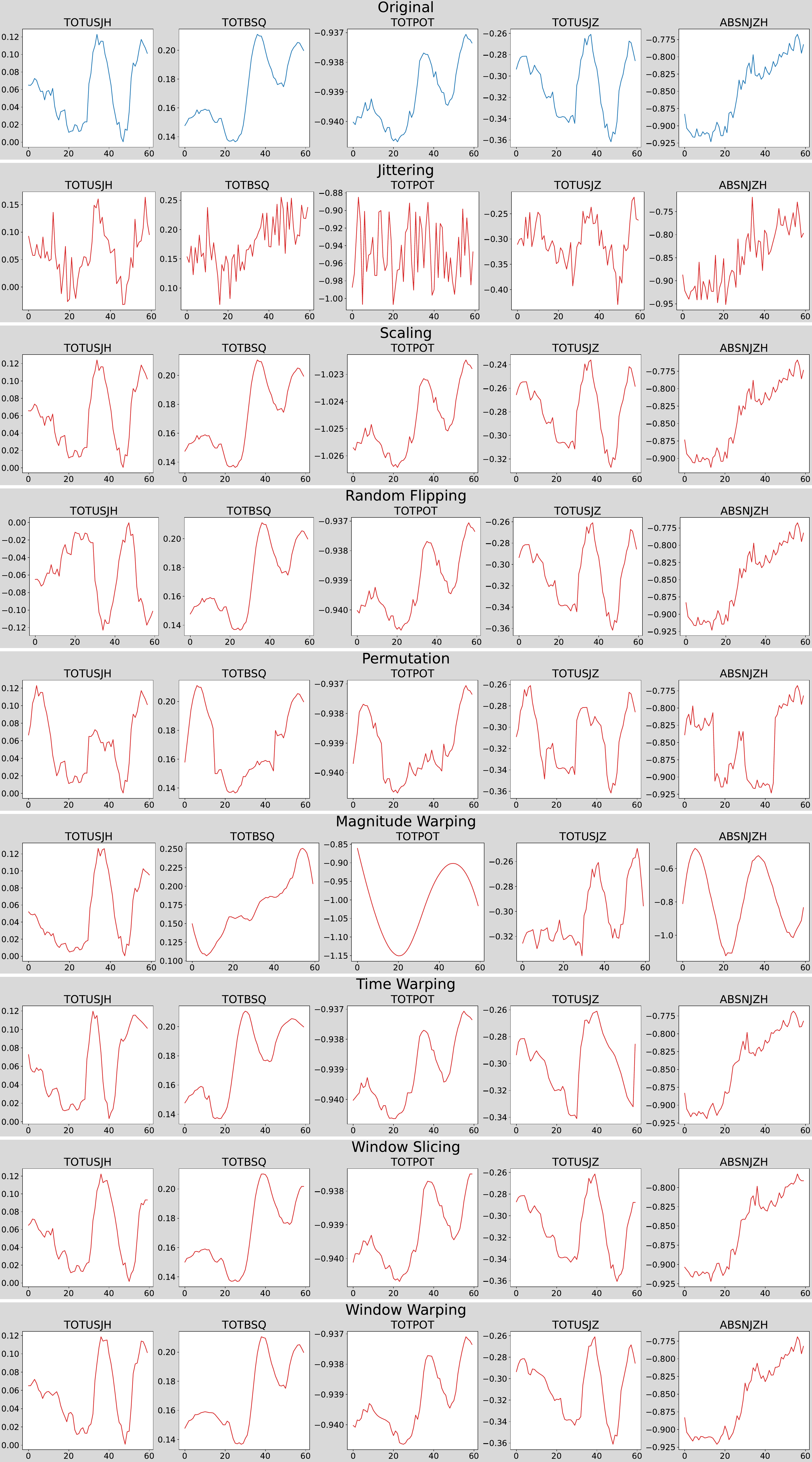}
        \caption{Examples of eight different basic data augmentation methods on a single MVTS data instance from the flaring class of SWAN-SF with five common features recommended in \cite{bobra2015solar}.}
        \label{fig:example}
    \end{figure*}

Most of the basic time series data augmentation methods are based on random transformations of training data and some of them are borrowed from image data augmentation. Given a MVTS data instance $T = \{t_1, \ldots, t_i, \ldots, t_n\} \in \mathbb{R}^{n \times D}$, where $n \in \mathbb{N}$ is the length of time series (i.e., number of time steps) and $D \in \mathbb{N}$ is the number of variate, a synthetic data instance $T'$ will be generated through a transformation function. Authors in \cite{iwana2021empirical} further divide those methods into three domains, the magnitude domain, time domain, and frequency domain. In this section, we give a description of eight different methods that we use in this study for multivariate time series (MVTS) data and their related works and an example of each method on SWAN-SF with five variables is shown in Fig. \ref{fig:example}.

\vspace{-3mm}
\subsubsection{Jittering}\label{section:jittering}
is the act of adding random noise to the values at each time step. It can be mathematically expressed as:
    \begin{equation}
        T' = \{t_1 + \epsilon_1, \ldots, t_i + \epsilon_i, \ldots, t_n + \epsilon_n\}, 
    \end{equation}
where $t_i$ is a vector containing the values at time step $i$, and $\epsilon_i$ is a vector of random noises generated from Gaussian distribution $\epsilon \sim \mathcal{N}(\mathbf{0},\,\mathbf{\sigma}^{2})$ with $\sigma$ being a hyperparameter that needs to be pre-determined. Jittering has been used for mitigating drift in time series data \cite{fields2019mitigating}. 

\vspace{-3mm}
\subsubsection{Scaling}\label{section:scaling}
for time series refers to transforming the magnitude of the time series by multiplying the values at each time step from a random scaling factor. Mathematically, scaling can be expressed as:. 
    \begin{equation}
        T' = \{\gamma \cdot t_1, \ldots, \gamma \cdot t_i, \ldots, \gamma \cdot t_n\}, 
    \end{equation}
where $\gamma$ is a vector of scaling factors generated from a Gaussian distribution $\gamma \sim \mathcal{N}(\mathbf{1},\,\mathbf{\sigma}^{2})$ with  $\mathbf{\sigma}$ a pre-determined hyperparameter. \cite{um2017data} used scaling to help with sensor data for Parkinson's disease monitoring. \cite{tran2020data} employed scaling as data augmentation for gait recognition.  


\vspace{-3mm}
\subsubsection{Rotation}\label{section:rotation}
in the context of multivariate time series involves randomly flipping and shuffling features to simulate rotational transformations. Mathematically, this can be expressed as:
    \begin{equation}
    T' = \{R \cdot t_1, \ldots, R \cdot t_i, \ldots, R \cdot t_n\},
    \end{equation}
where $R$ is a vector of random rotation angles, each element generated from a uniform distribution, $R \sim U(\mathbf{-1},,\mathbf{1})$. While some studies like \cite{um2017data} have observed improvements in accuracy when rotation is used, particularly when combined with other data augmentation techniques, due to the challenge of interpreting the shuffling of time series across different features, our study opts for a simpler approach termed ``random flipping", where we flip the time series of randomly selected features only.

\vspace{-3mm}
\subsubsection{Magnitude Warping}\label{section:magnitude_warping}
changes the magnitude of $T$ by convolving the data window with a smooth curve varying around one \cite{um2017data}. Mathematically, magnitude warping can be defined as:
    \begin{equation}
        T' = \{\gamma_1 \cdot t_1, \ldots, \gamma_i \cdot t_i, \ldots, \gamma_n \cdot t_n\}, 
    \end{equation}
where $\{\gamma_1, \ldots, \gamma_i, \ldots, \gamma_n\}$ is a matrix generated by interpolation of a cubic spline $S(\mathbf{x})$ with knots $\mathbf{x} \in \mathbb{R}^{k}$ being generated from a Gaussian distribution $\mathcal{N}(\mathbf{1},\,\mathbf{\sigma}^{2})$. The number of knots $k$ and the standard deviation $\sigma$ are the hyperparameters to be pre-determined.

\vspace{-3mm}
\subsubsection{Slicing}\label{section:window_slicing}
is the idea of cutting off a window of size $W$ from the original time series. Mathematically, slicing can be defined as:
    \begin{equation}
        T' = \{t_\varphi, \ldots, t_i, \ldots, t_{W+\varphi}\}, 
    \end{equation}
where $\varphi$ is an integer that is randomly selected such that $1 \leq \varphi \leq n-W$. In this way, slicing is equivalent to the operation named Window Slicing (WS) in \cite{le2016data}.

\vspace{-3mm}
\subsubsection{Permutation}\label{section:permutation}
involves rearranging segments of a time series to generate novel patterns \cite{um2017data}. This method can be applied by either equally sized segments or segments of variable sizes. Permutation with equally sized segments will split the time series into $N$ segments with a length of $\frac{T}{N}$ and then permutes them, while permutation with variable sized of segments will split the time series into random sizes. Moreover, permutation does not maintain the time dependencies present in the original time series. 

\vspace{-3mm}
\subsubsection{Time Warping}\label{section:time_warping}
refers to the manipulation of temporal patterns, achieved either through a smooth warping path \cite{um2017data} or a randomly positioned fixed window \cite{le2016data}. In the context of a smooth warping path, time warping can be mathematically expressed as:
    \begin{equation}
        T' = \{t_{\tau(1)}, \ldots, t_{\tau(i)}, \ldots, t_{\tau(n)}\}, 
    \end{equation}
where $\tau(\cdot)$ represents a warping function. This function adjusts the time steps based on a smooth curve defined by a cubic spline $S(\mathbf{x})$ with each knot $x_i$ of knots $\mathbf{x} \in \mathbb{R}^{k}$ sampled from a normal distribution $\mathcal{N}(1,\,\mathbf{\sigma}^{2})$.

\vspace{-3mm}
\subsubsection{Window Warping}\label{section:window_warping}
is a variant of time warping proposed in \cite{le2016data}. It offers a distinct method for manipulating temporal patterns within a time series. This technique involves selecting a random window from the time series and altering its duration by either stretching it twofold or contracting it by half. Although initially set at fixed values of $\frac{1}{2}$ and 2, the authors highlight the adaptability of these multipliers, indicating they can be customized or fine-tuned to alternative values depending on the desired outcome or application.

\subsection{SWAN-SF Dataset}\label{section:swan-sf}
Space Weather ANalytics for Solar Flares (SWAN-SF) is a multivariate time series (MVTS) dataset for benchmarking solar flare prediction introduced in \cite{angryk2020multivariate}. The dataset covers 4,098 MVTS data collections of over eight years of solar activities from Solar Cycle 24. Each MVTS data instance represents a 12-hour observation window of 51 flare-predictive parameters. Within the observation window, each time series is collected with the full 12-minute dance of the Solar Dynamics Observatory (SDO)/Helioseismic and Magnetic Imager (HMI) definitive vector magnetogram data, resulting in 5 values per hour. Consequently, each observation-window time series for each active region comprises 60 data values. An MVTS data instance is labeled by the class of the strongest flare reported within a 24-hour prediction window right after the observation window (namely, with zero latency). Based on the peak soft X-ray flux observed by the X-ray sensor onboard the Geostationary Operational Environmental Satellite (GOES), solar flares can be categorized into five classes denoted by letters A, B, C, M, and X, from weakest to the strongest. In SWAN-SF, instances with no-flares or only A-class flares in the prediction window are labeled by N.   

The SWAN-SF dataset is created through a sliding-window methodology with a 1-hour step size, leading to temporal coherence issues that can introduce bias and overfitting during model training \cite{ahmadzadeh2021train}. To address this, SWAN-SF is divided into five partitions without temporal overlapping with each partition containing approximately equal number of M- and X-class instances. Additionally, SWAN-SF is highly imbalanced where M- and X-class instances are extremely less than the instances of the remaining three classes. Tab. \ref{tab:swan-sf} shows the number of instances for each class and the imbalance ratio in each partition. Thus, SWAN-SF serves as a pertinent dataset for studying rare-event prediction, characterized by limited data availability and class imbalance, making it well-suited for this study.
    \begin{table*}[!tp]
    \caption{Sample sizes and imbalance ratios in each partition in SWAN-SF. The time spans between two adjacent partitions are not overlapping.}
        \centering
        \begin{tabular}{c@{\hspace{4mm}}c@{\hspace{4mm}}cccccD{:}{:}{4.6}D{:}{:}{4.5}}
        \hline \hline
        \multirow{2}{*}{\textbf{Partition}} & \multirow{2}{*}{\textbf{Time Span}} & \multicolumn{5}{c}{\textbf{Class Distribution}} & \multicolumn{1}{c}{\textbf{{\hspace{2mm}}Imbalance Ratio}} \\
        \cline{3-7}
         & & X & M & C & B & N &  {\hspace{2mm}}(\textup{MX}:\textup{NBC}) \\
        \hline
        1	& 05/2010 – 03/2012	& 165 & 1,089 & 6,416 & 5,692 & 60,130 & {\hspace{2mm}}1:58\\
        2	& 03/2012 – 10/2013	& 72 & 1,392 & 8,810 & 4,978 & 73,368 & {\hspace{2mm}}1:62\\
        3	& 10/2013 – 03/2014	& 136 & 1,288 & 5,639 & 685 & 34,762 & {\hspace{2mm}}1:29\\
        4	& 03/2014 – 03/2015	& 153 & 1,012 & 5,956 & 846 & 43,294 & {\hspace{2mm}}1:43\\
        5	& 03/2015 – 08/2018	& 19 & 971 & 5,753 & 5,924 & 62,688 & {\hspace{2mm}}1:75\\
        \hline
        \end{tabular}
        \label{tab:swan-sf}
    \vspace{-3mm}
    \end{table*}

\section{Mean Gaussian Noise for Multivariate and Underrepresented Time Series Data}\label{section:mgn}
\begin{figure*}
        \centering
        \includegraphics[width=0.9\textwidth,keepaspectratio]{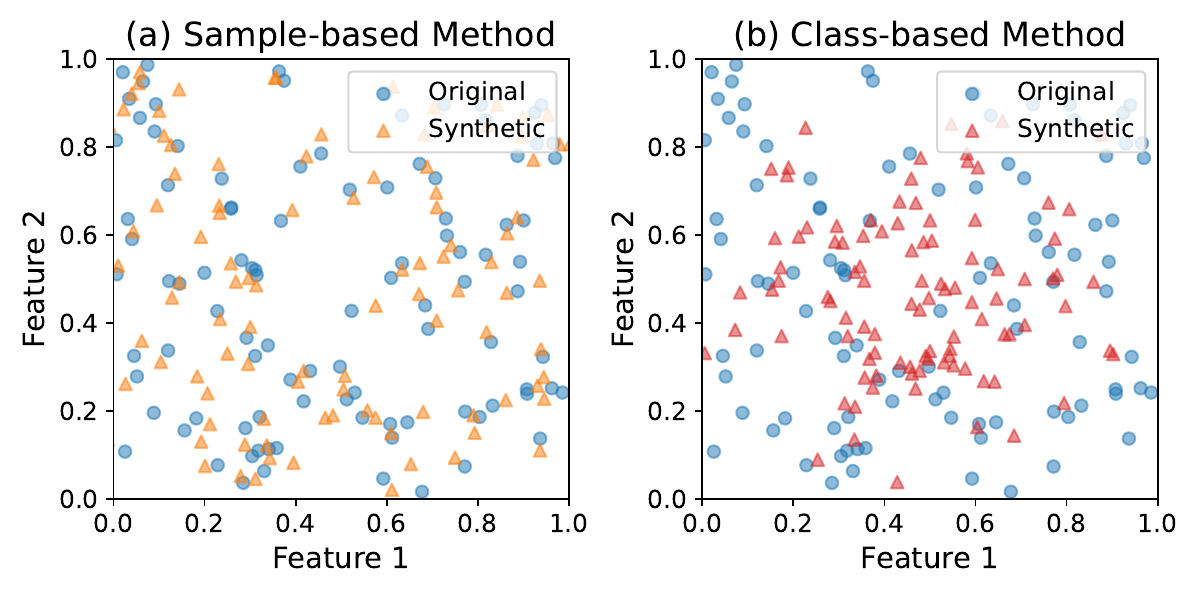}
        \vspace{-5mm}
        \caption{The difference in data generation between sample-based methods and class-based methods demonstrated using randomly generated 2D data. The synthetic data in (a) is generated by jittering and the synthetic data in (b) is generated by MGN.}
        \label{fig:sample-based_vs_class-based}
        \vspace{-5mm}
    \end{figure*}

In this section, we introduce the novel data augmentation technique called Mean Gaussian Noise (MGN), which diverges from conventional augmentation techniques discussed in Section \ref{section:basic_augmentation}. While traditional methods focus on generating new/synthetic data through manipulating individual time series data instances within a dataset, MGN takes a global approach by generating synthetic data using the statistical characteristics of the entire input dataset (i.e., the class to be augmented). This class-based method aims to improve the representation of central tendency/trend of the underrepresented classes in the input space, while \text{maintaining the original seperation} of classes at their boarders. The difference between traditional sample-based augmentation methods (e.g., jittering) and our class-based method (i.e., MGN) is illustrated in Fig. \ref{fig:sample-based_vs_class-based}.

The fundamental concept behind MGN is to introduce Gaussian noise centered around the mean values of the input multivariate time series dataset. By capturing the overall statistical tendencies of the data, MGN aims to create instances that reflect the underlying distribution of the underrepresented data. This global perspective allows MGN to generate samples that are representative of the rare class as a whole, rather than being limited to specific instances.

\begin{figure*}[!thp]
        \centering
        \includegraphics[width=\textwidth,keepaspectratio]{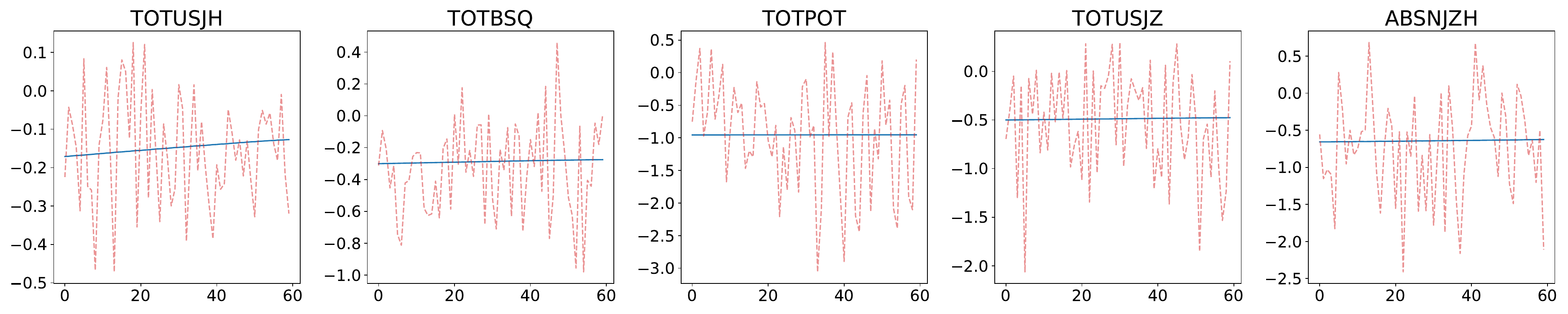}
        \vspace{-5mm}
        \caption{Examples of Mean Gaussian Noise (MGN) applied to the common five features \cite{bobra2015solar} of the flaring data (i.e., extremely rare class) in Partition 1 from SWAN-SF. The blue lines represent the mean time series and the dotted red lines are the generated time series by MGN. The values on the y-axis for each feature are normalized to compare with other methods (e.g., Fig. \ref{fig:example}).}
        \label{fig:example_mgn}
        \vspace{-5mm}
    \end{figure*}

Initially, MGN computes the mean values along each feature at each time step for the whole population of the input dataset, effectively summarizing the central tendencies of the data across different features and time steps. These mean values serve as reference points for the generation of synthetic instances. Subsequently, random Gaussian noise is added to each mean value at every time step, introducing variability while preserving the overall structure of the class. Formally, MGN can be expressed as:
    \begin{equation}
        T' = \{\bar{t_1} \cdot (\mathbf{1} + \epsilon_1), \ldots, \bar{t_i} \cdot (\mathbf{1} + \epsilon_i), \ldots, \bar{t_n} \cdot (\mathbf{1} + \epsilon_n)\}, 
    \end{equation}
where $\bar{t_i} \in \mathbb{R}^D$ represents a vector that contains the mean values of $D$ individual descriptive features at time step $i$ and is calculated as $\frac{1}{n} \sum_{k=1}^{n} t_{ki}$ with $t_{ki}$ being the value at time step $i$ of $k$-th instance. The $\epsilon_i \in \mathbb{R}^D$ denotes a vector of random noises, each of which is generated from a Gaussian distribution $\mathcal{N}(0,\,\mathbf{\sigma}^{2})$ with standard deviation $\sigma$ being a parameter that controls the magnitude of noises to be added, offering flexibility in the augmentation process. An example of MGN-generated results for five physical parameters (i.e., descriptive features) from SWAN-SF dataset is shown in Fig. \ref{fig:example_mgn}.

    \vspace{-3mm}

While jittering also generates synthetic data by adding Gaussian noise, it operates directly on individual time series data instances. However, this approach may limit the variability of data generation, potentially constraining exploration in the input space and adding overlapped samples at the boarder between rare and frequent classes. In contrast, MGN introduces more flexibility by allowing greater randomness in data generation while preserving the overall underlying distribution of the input dataset. Moreover, MGN adds noise proportionally along with each feature dimension through $\bar{t_1} \cdot (\mathbf{1} + \epsilon_1)$, which offers more control over the scale and magnitude of the noise. We anticipate that MGN will enhance coverage of the input space when mapping to the high-dimensional representations for ML training.

Similar to existing methods like jittering and scaling, MGN requires only one parameter, the standard deviation $\sigma$ of the Gaussian noise, to be configured. However, MGN's approach is advantageous for multivariate time series as it does not necessitate feature normalization before augmentation. This flexibility makes MGN a valuable addition to the toolkit of data augmentation techniques for multivariate time series datasets.

\section{Experiment Setup}\label{section:exp_setup}
In this study, we focus on investigating effectiveness of data augmentation techniques for a task of binary classification for solar flare prediction using SWAN-SF. Among the five classes in SWAN-SF, we categorize M and X as positive class (referred to as MX) due to the heightened risk they pose to human society. Conversely, the remaining three classes, N, B, and C, are grouped into negative class (referred to as NBC). The actual imbalance ratios in our experiments are shown in Tab. \ref{tab:swan-sf}. To mitigate the curse of dimensionality, we focus on the top five predictive SHARP parameters suggested in \cite{bobra2015solar}: total unsigned current heilicy (TOTUSJH), total magnitude of Lorentz force (TOTBSQ), total photospheric magnetic free energy density (TOTPOT), total unsigned vertical current (TOTUSJZ), and  absolute value of the net current helicity (ABSNJZH). Furthermore, we maintain consistency for different augmentation methods by following the hyperparameter settings used in \cite{iwana2021empirical}. The standard deviation $\sigma$ for the MGN method is set to 1.0. 

Leveraging the fact that SWAN-SF is split into five partitions without temporal overlapping, we use Partition 1 for training and Partitions 2-5 for testing. For each data augmentation method detailed in Section \ref{section:basic_augmentation}, we generate one synthetic instance for each instance in MX class and combine them with the original data, which doubles the training size. To ensure consistency in training data size across all experiments, for our baseline, we duplicate the MX class once to match the augmented data size. For MGN, we specify the number of samples to generate as 1,254 (i.e., the count of MX class in Partition 1).

Random undersampling without replacement is applied to the NBC class, where an equal number of NBC-class instances are randomly selected to balance the training data. Consequently, the total training data comprises 2,508 instances for both the MX and NBC classes accross all our experiments. To ensure the reliability of our experiments, we repeat the random undersampling without replacement process ten times, each time selecting different instances from NBC class. We utilize the mean and standard deviation of the evaluation metrics from the ten runs to assess the final model performance. 

The binary classifier employed in our experiment is TimeSeriesSVC from tslearn library \cite{tavenard2020tslearn}. Tslearn is a Python machine learning tool specifically designed for time series data and is build upon the Scikit-learn library \cite{pedregosa2011scikit}, Numpy \cite{harris2020array}, and SciPy \cite{virtanen2020scipy}. TimeSeriesSVC extends the traditional Support Vector Machine (SVM) algorithm to effectively handle temporal dependencies and patterns inherent in time series datasets. Similar to conventional SVM, TimeSeriesSVC takes three hyperparameters: kernel function, kernel coefficient $\gamma$, and soft margin constant $C$. In our study, to ensure a fair evaluation, we consistently use the radial basis function (RBF) kernel with $\gamma=0.01$ and $C=1$ to train the classifier across all data augmentation methods. 

Given the imbalanced nature of the testing data, traditional evaluation metrics such as accuracy may provide misleading insights \cite{ahmadzadeh2021train}. To ensure a comprehensive evaluation of model performance, we rely on two widely-used \cite{wen2022improving, chen2022cgan} metrics in the space weather community: the true skill score (TSS) \cite{bloomfield2012toward} and the updated Heidke skill score (HSS2) \cite{bobra2015solar}. TSS is calculated as:
    \begin{equation}
    \label{equ:tss}
    TSS = \frac{TP}{TP+FN} - \frac{FP}{FP+TN},
    \end{equation}  
where TP, FP, TN, and FN represent true positives, false positives, true negatives, and false negatives, respectively. TSS measures the difference between the true positive rate (recall) and the false alarm rate, with values ranging from -1 to 1. A TSS score of 1 indicates perfect performance, while -1 suggests that all predictions made by the classifier are reversed. 

On the other hand, HSS2 emphasizes the model's ability to make correct positive predictions (TP) while minimizing false alarms (FP), and is calculated as:
    \begin{equation}
    \label{equ:hss2}
    HSS2 = \frac{2(TP\cdot TN - FN\cdot FP)}{P(FN+TN)+N(TP+FP)},
    \end{equation}
where P and N represent the total number of positive and negative instances, respectively (each 2,508 in all our experiments). Similar to TSS, HSS2 scores range from -1 to 1, with 1 indicating a perfect model and -1 implying a reversal of labels for all testing instances. By combining TSS and HSS2, we obtain a comprehensive assessment of the model's performance on imbalanced data, considering both discrimination and reliability.

\section{Results and Discussion}
\label{section:exp_result}
    \begin{table*}[t]
    \caption{DtP of different data augmentation methods on each testing partition. Each cell shows the mean and standard deviation (in the parenthesis) across ten runs of our random undersampling experiments.}
        \centering
        \begin{tabular}{l | c | c | c | c }
        \hline
        \textbf{Augmentation Method} &  \textbf{Partition 2} & \textbf{Partition 3} & \textbf{Partition 4} & \textbf{Partition 5}\\
        \Xhline{3\arrayrulewidth}
        Original/Baseline & 0.852 (0.004) & 0.790 (0.003) & 0.796 (0.010) & 0.825 (0.005) \\ \hline
        Jittering & 0.850 (0.003) & 0.793 (0.011) & 0.791 (0.002) & 0.823 (0.004) \\ \hline
        Scaling & 0.847 (0.006) & 0.790 (0.012) & 0.789 (0.004) & 0.821 (0.006) \\ \hline
        Random Flipping & 0.800 (0.007) & \textbf{0.761} (0.007) & 0.743 (0.009) & 0.778 (0.004) \\ \hline
        Permutation & 0.853 (0.004) & 0.793 (0.003) & 0.797 (0.004) & 0.829 (0.004) \\ \hline
        Magnitude Warping & 0.846 (0.010) & 0.796 (0.024) & 0.827 (0.050) & 0.817 (0.008) \\ \hline
        Time Warping & 0.852 (0.004) & 0.791 (0.003) & 0.794 (0.003) & 0.826 (0.005) \\ \hline
        Slicing & 0.852 (0.004) & 0.791 (0.002) & 0.797 (0.010) & 0.826 (0.005) \\ \hline 
        Window Warping & 0.852 (0.004) & 0.790 (0.003) & 0.797 (0.011) & 0.825 (0.005) \\ \hline 
        Mean Gaussian Noise & \textbf{0.767} (0.029) & 0.763 (0.019) & \textbf{0.719} (0.041) & \textbf{0.743} (0.027) \\
        \hline
        \end{tabular}
        \label{tab:exp_res}
    \vspace{-3mm}
    \end{table*}
    
\begin{figure*}[!h]
    \vspace{-3mm}
        \centering
        \includegraphics[width=0.9\textwidth,keepaspectratio]{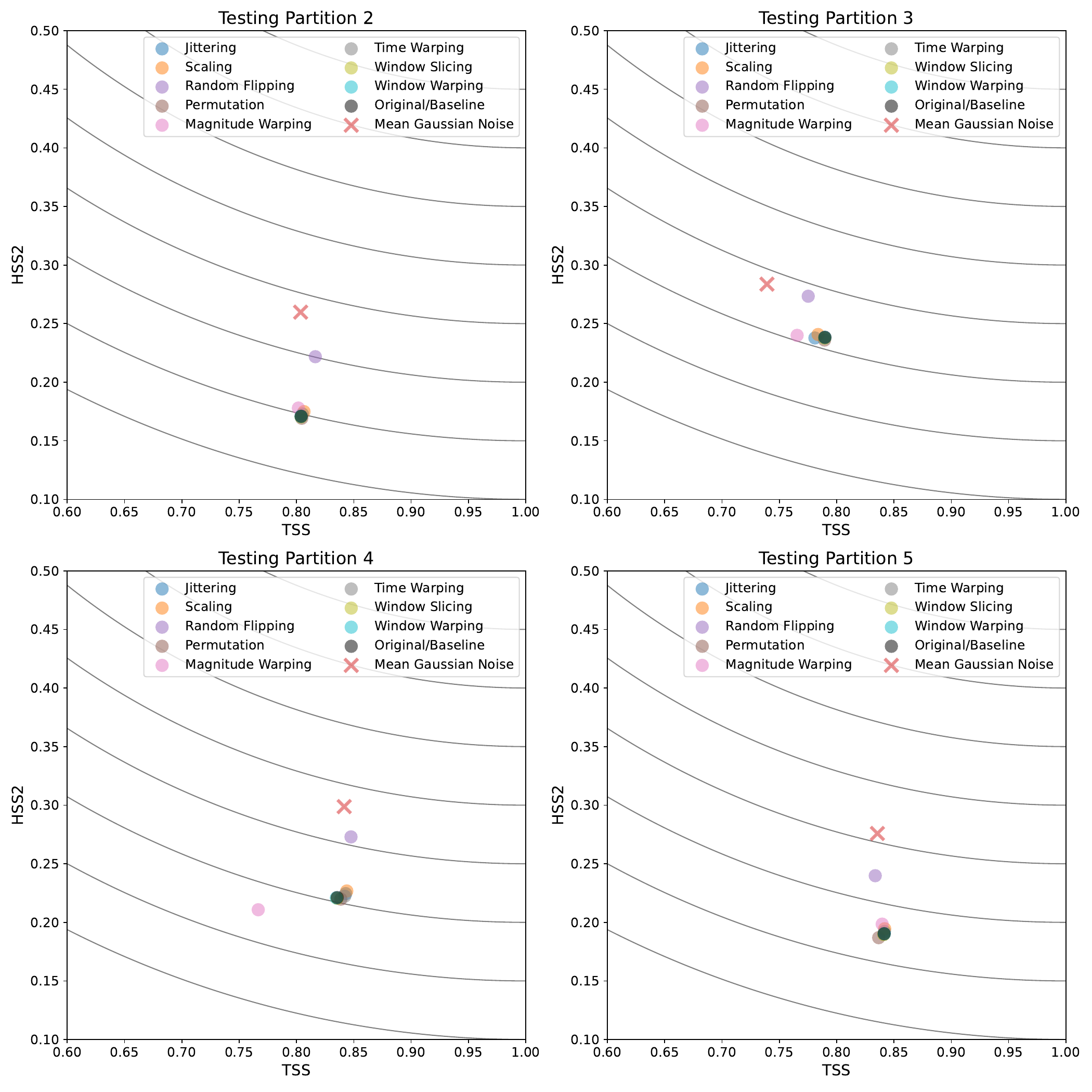}
        \vspace{-5mm}
        \caption{The result of comparison between different data augmentation methods. Each dot represents the mean values of TSS and HSS2 of the ten runs of random undersampling for the corresponding data augmentation method. Both axes are zoomed in to enhance detail and improve visualization.}
        \label{fig:exp_result}
    \vspace{-3mm}
    \end{figure*}

To visually compare the performance of our models based on the two evaluation metrics, we plot the true skill score (TSS) against the updated Heidke skill score (HSS2) for each testing partition 2-5 in Fig. \ref{fig:exp_result}. In each plot, moving toward the top right corner reflects improving prediction performance, any points lying on the same curve have a same distance from the coordinate (1,1) and are considered to have equivalent performance levels. From Fig. \ref{fig:exp_result}, we observe that MGN and random flipping exhibit notably superior performance compared to other evaluated augmentation techniques, while the remaining methods exhibit similar performance to the baseline across all testing partitions. To quantitatively assess the distance of each point from the perfect performance corner, we compute the Euclidean distance, termed distance to the perfect (DtP), using the formula:
    \begin{equation}
        DtP = \sqrt{(1-TSS)^2 + (1-HSS2)^2}.
    \end{equation}

The DtP metric ranges from 0 to $2\sqrt{2}$, where 0 denotes a perfect model and $2\sqrt{2}$ signifies a classifier that assigns the opposite labels to all testing instances. Tab.\ref{tab:exp_res} shows the mean and standard deviation (in the parentheses) of DtP for different augmentation methods across ten runs of random undersampling. As illustrated, MGN achieves the best performance on three testing partitions, while preserving competitive performance with random flipping on the remaining partition.

The primary concept of data augmentation is to generate synthetic data to explore unexplored regions of the input space. However, most augmentation methods tested in our study manipulate individual instances, potentially limiting exploration in high-dimensional spaces. Random flipping, in contrast, flips the time series of randomly selected features, which may help the generated data cover more input space. MGN, from a different persepective, generates data globally using the mean of the entire dataset's time series, which may stimulate the exploration in the input space as well. That may also be the reason why MGN has greater variances than the other methods. 

\section{Run Time Experiments}
\label{section:time}
\begin{figure*}[!th]
    \vspace{-3mm}
        \centering
        \includegraphics[width=0.95\textwidth,keepaspectratio]{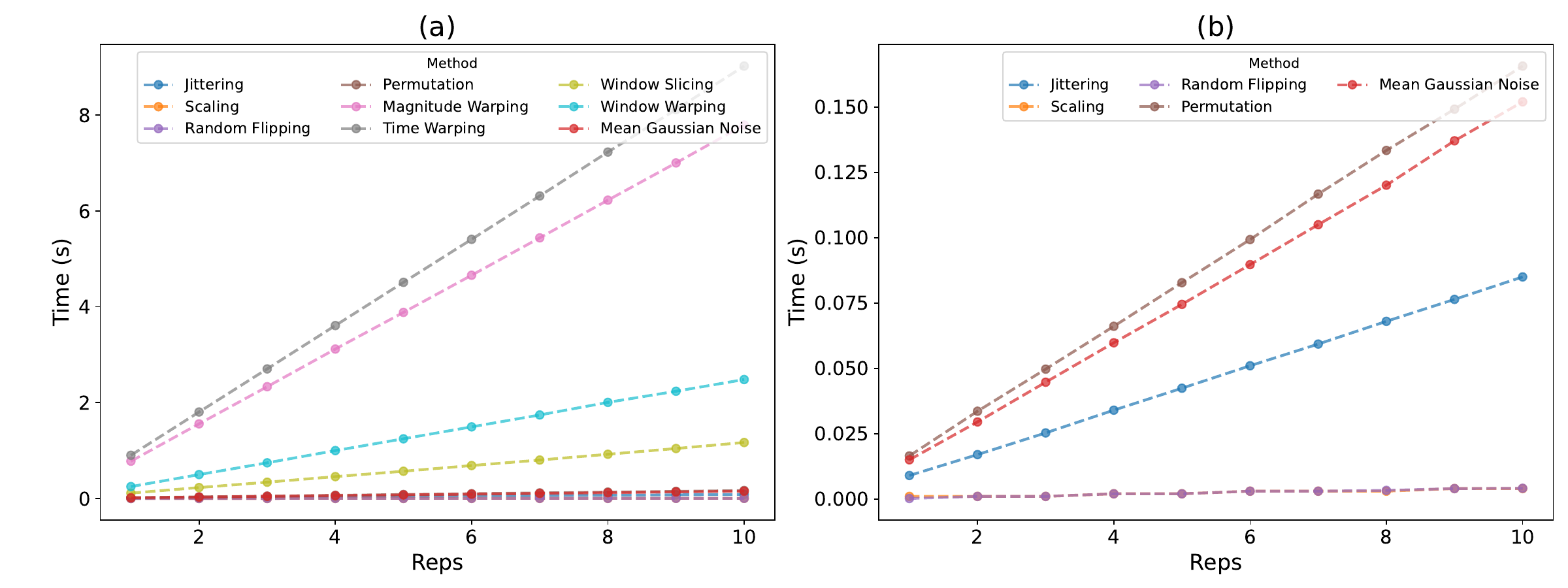}
        \vspace{-5mm}
        \caption{Run time of different data augmentation methods. (a) shows the computational cost of all nine different augmentation methods evaluated in this study. (b) provides a close-up view of the five overlapping methods from (a) for enhanced clarity.}
        \label{fig:time}
    \vspace{-3mm}
    \end{figure*}

In this section, we conduct an experimental time complexity investigation on various augmentation methods. We generate different numbers of data for each method based on the number of multiplications of the MX-class instances in Partition 1 from SWAN-SF with the five parameters mentioned in Section \ref{section:exp_setup}. We use a variable named \textit{Reps} to represent the number of repetitions of MX-class instances to generate. For instance, when $Reps=1$, 1254 synthetic instances will be generated. We test a set of \textit{Reps} from 1 to 10 and run ten times for each value. Subsequently, we take the average as the final running time for each method. As depicted in Fig. \ref{fig:time}, MGN emerges as competitive in the regards to computational cost while yielding the best performance in terms of DtP.

\section{Conclusion and Future Work}
\label{section:conclusion}
In this study, we introduce a novel approach termed Mean Gaussian Noise (MGN), which generates synthetic data globally by utilizing the mean of the entire dataset's time series and involves only one pre-determined hyperparameter. We conduct an experiment to compare MGN with various basic data augmentation methods on a multivariate time series dataset, SWAN-SF, for solar flare prediction. Our experimental results show that MGN exhibits better performance compared to other investigated methods. This may be attributed to the increased randomness inherent in the method, which facilitates exploration of the input space. Moreover, our run time experiments reveal that MGN operate more efficiently than certain methods and remain competitive with the others. Moving forward, our future research endeavors aim to extend the evaluation of our methods across a broader range of multivariate time series datasets and with diverse classifiers, including neural networks. Additionally, we aspire to develop more advanced data augmentation techniques capable of generating even more accurate synthetic data, with the goal of improving machine learning for extremely imbalanced data.

\vspace{-2mm}
\subsubsection{\ackname} This project is supported in part by funding from CISE, MPS and GEO Directorates under NSF award \verb|#|1931555, and by funding from NASA, under awards \verb|#|80NSSC23K1026 and \verb|#|80NSSC24K0238.
        
%
%
%
\bibliographystyle{splncs04}
\bibliography{mybib}

\begin{thebibliography}{10}
\providecommand{\url}[1]{\texttt{#1}}
\providecommand{\urlprefix}{URL }
\providecommand{\doi}[1]{https://doi.org/#1}

\bibitem{ahmadzadeh2021train}
Ahmadzadeh, A., Aydin, B., Georgoulis, M.K., Kempton, D.J., Mahajan, S.S., Angryk, R.A.: How to train your flare prediction model: Revisiting robust sampling of rare events. The Astrophysical Journal Supplement Series  \textbf{254}(2), ~23 (2021)

\bibitem{angryk2020multivariate}
Angryk, R.A., Martens, P.C., Aydin, B., Kempton, D., Mahajan, S.S., Basodi, S., Ahmadzadeh, A., Cai, X., Filali~Boubrahimi, S., Hamdi, S.M., et~al.: Multivariate time series dataset for space weather data analytics. Scientific data  \textbf{7}(1), ~227 (2020)

\bibitem{batista2004study}
Batista, G.E., Prati, R.C., Monard, M.C.: A study of the behavior of several methods for balancing machine learning training data. ACM SIGKDD explorations newsletter  \textbf{6}(1),  20--29 (2004)

\bibitem{bloomfield2012toward}
Bloomfield, D.S., Higgins, P.A., McAteer, R.J., Gallagher, P.T.: Toward reliable benchmarking of solar flare forecasting methods. The Astrophysical Journal Letters  \textbf{747}(2), ~L41 (2012)

\bibitem{bobra2015solar}
Bobra, M.G., Couvidat, S.: Solar flare prediction using sdo/hmi vector magnetic field data with a machine-learning algorithm. The Astrophysical Journal  \textbf{798}(2), ~135 (2015)

\bibitem{chen2022cgan}
Chen, Y., Kempton, D.J., Ahmadzadeh, A., Wen, J., Ji, A., Angryk, R.A.: Cgan-based synthetic multivariate time-series generation: a solution to data scarcity in solar flare forecasting. Neural Computing and Applications  \textbf{34}(16),  13339--13353 (2022)

\bibitem{fields2019mitigating}
Fields, T., Hsieh, G., Chenou, J.: Mitigating drift in time series data with noise augmentation. In: 2019 International Conference on Computational Science and Computational Intelligence (CSCI). pp. 227--230. IEEE (2019)

\bibitem{gao2023interpretable}
Gao, Y., Lewis, N., Calhoun, V.D., Miller, R.L.: Interpretable lstm model reveals transiently-realized patterns of dynamic brain connectivity that predict patient deterioration or recovery from very mild cognitive impairment. Computers in Biology and Medicine  \textbf{161},  107005 (2023)

\bibitem{harris2020array}
Harris, C.R., Millman, K.J., Van Der~Walt, S.J., Gommers, R., Virtanen, P., Cournapeau, D., Wieser, E., Taylor, J., Berg, S., Smith, N.J., et~al.: Array programming with numpy. Nature  \textbf{585}(7825),  357--362 (2020)

\bibitem{iwana2021empirical}
Iwana, B.K., Uchida, S.: An empirical survey of data augmentation for time series classification with neural networks. Plos one  \textbf{16}(7),  e0254841 (2021)

\bibitem{le2016data}
Le~Guennec, A., Malinowski, S., Tavenard, R.: Data augmentation for time series classification using convolutional neural networks. In: ECML/PKDD workshop on advanced analytics and learning on temporal data (2016)

\bibitem{mehdiyev2017time}
Mehdiyev, N., Lahann, J., Emrich, A., Enke, D., Fettke, P., Loos, P.: Time series classification using deep learning for process planning: A case from the process industry. Procedia Computer Science  \textbf{114},  242--249 (2017)

\bibitem{mudelsee2019trend}
Mudelsee, M.: Trend analysis of climate time series: A review of methods. Earth-science reviews  \textbf{190},  310--322 (2019)

\bibitem{pedregosa2011scikit}
Pedregosa, F., Varoquaux, G., Gramfort, A., Michel, V., Thirion, B., Grisel, O., Blondel, M., Prettenhofer, P., Weiss, R., Dubourg, V., et~al.: Scikit-learn: Machine learning in python. the Journal of machine Learning research  \textbf{12},  2825--2830 (2011)

\bibitem{shorten2019survey}
Shorten, C., Khoshgoftaar, T.M.: A survey on image data augmentation for deep learning. Journal of big data  \textbf{6}(1),  1--48 (2019)

\bibitem{shorten2021text}
Shorten, C., Khoshgoftaar, T.M., Furht, B.: Text data augmentation for deep learning. Journal of big Data  \textbf{8}(1), ~101 (2021)

\bibitem{tavenard2020tslearn}
Tavenard, R., Faouzi, J., Vandewiele, G., Divo, F., Androz, G., Holtz, C., Payne, M., Yurchak, R., Ru{\ss}wurm, M., Kolar, K., et~al.: Tslearn, a machine learning toolkit for time series data. Journal of machine learning research  \textbf{21}(118), ~1--6 (2020)

\bibitem{tran2020data}
Tran, L., Choi, D.: Data augmentation for inertial sensor-based gait deep neural network. IEEE Access  \textbf{8},  12364--12378 (2020)

\bibitem{tsay2005analysis}
Tsay, R.S.: Analysis of financial time series. John wiley \& sons (2005)

\bibitem{um2017data}
Um, T.T., Pfister, F.M., Pichler, D., Endo, S., Lang, M., Hirche, S., Fietzek, U., Kuli{\'c}, D.: Data augmentation of wearable sensor data for parkinson’s disease monitoring using convolutional neural networks. In: Proceedings of the 19th ACM international conference on multimodal interaction. pp. 216--220 (2017)

\bibitem{virtanen2020scipy}
Virtanen, P., Gommers, R., Oliphant, T.E., Haberland, M., Reddy, T., Cournapeau, D., Burovski, E., Peterson, P., Weckesser, W., Bright, J., et~al.: Scipy 1.0: fundamental algorithms for scientific computing in python. Nature methods  \textbf{17}(3),  261--272 (2020)

\bibitem{wen2022improving}
Wen, J., Islam, M.R., Ahmadzadeh, A., Angryk, R.A.: Improving solar flare prediction by time series outlier detection. In: International Conference on Artificial Intelligence and Soft Computing. pp. 152--164. Springer (2022)

\end{thebibliography}
%




\end{document}